\documentclass[letterpaper, 10 pt, conference]{ieeeconf}  

\IEEEoverridecommandlockouts                              
\usepackage[utf8]{inputenc}
\usepackage[T1]{fontenc}

\usepackage{graphicx} 
\usepackage{amsmath} 
\usepackage{amssymb}  
\usepackage{algorithm,algorithmic}
\usepackage{array}
\usepackage[hyphens]{url}

\title{\LARGE \bf
Sample-efficient Reinforcement Learning in Robotic Table Tennis*
}

\author{Jonas Tebbe$^{1}$, Lukas Krauch$^{1}$, Yapeng Gao$^{1}$, and Andreas Zell$^{1}$
\thanks{*This work was supported in parts by the Vector Stiftung and KUKA.}
\thanks{$^{1}$Jonas Tebbe, Lukas Krauch, Yapeng Gao and Andreas Zell are with the Cognitive Systems group, Computer Science Department,
        University of Tuebingen, 72076 Tuebingen, Germany
        {\tt\small [jonas.tebbe, yapeng.gao, andreas.zell]@uni-tuebingen.de, lukas.krauch@student.uni-tuebingen.de}}%
}

\begin{document}

\maketitle
\thispagestyle{empty}
\pagestyle{empty}

\begin{abstract}

Reinforcement learning (RL) has achieved some impressive recent successes in various computer games and simulations. Most of these successes are based on having large numbers of episodes from which the agent can learn. In typical robotic applications, however, the number of feasible attempts is very limited. In this paper we present a sample-efficient RL algorithm applied to the example of a table tennis robot. In table tennis every stroke is different, with varying placement, speed and spin. An accurate return therefore has to be found depending on a high-dimensional continuous state space. To make learning in few trials possible the method is embedded into our robot system. In this way we can use a one-step environment. The state space depends on the ball at hitting time (position, velocity, spin) and the action is the racket state (orientation, velocity) at hitting. An actor-critic based deterministic policy gradient algorithm was developed for accelerated learning. Our approach performs competitively both in a simulation and on the real robot in a number of challenging scenarios. Accurate results are obtained without pre-training in under $200$ episodes of training. The video presenting our experiments is available at \textnormal{\url{https://youtu.be/uRAtdoL6Wpw}}.
\end{abstract}

\section{Introduction}
Reinforcement learning (RL) is, next to supervised and unsupervised learning, one of the three basic machine learning areas. RL is a technique in which an artificial agent or a robot learns an optimal decision-making policy in a specific environment by trial and error. In recent times, reinforcement learning has come to great success in various video  and board games such as Atari-Games \cite{Rainbow,Agent57} and Go \cite{AlphaZero}. After OpenAI introduced new robotic environments \cite{OpenAiGym,RoboticGym}, strong results were also achieved for simulations of various robotic scenarios \cite{DDPG,HER,SAC}. 

This suggests that these learning algorithms might also be used to control real robots. It could shorten the development process for new control algorithms and thus bring robotics into other previously unavailable application areas. However, it is not possible to adapt these successful methods directly \cite{rlblogpost}. Millions of attempts are often required to solve a task such as playing an Atari game. On a real robot this is not feasible in a reasonable amount of time. In addition, exhaustive exploration strategies are often not suitable without damaging the robot and its environment.  

In this paper we present a reinforcement learning algorithm for a table tennis playing robot, which addresses two key problems of realistic reinforcement learning applications in robotics. The first one is sample efficiency. In every scenario learning from scratch is possible with only a small dataset of fewer than $200$ ball returns. Secondly, robustness is essential as the robot is facing multiple sources of real world noise in ball measurements, trajectory predictions and arm movements. Our key contributions are summarized by the following:
\begin{itemize}
\item A simulation was developed and used to tune an RL algorithm based on DDPG for rapid learning on small datasets. 

\item The RL algorithm was integrated into a real robotic table tennis system. 

\item Extensive experiments were conducted to access the performance of the method for a range of scenarios with varying degrees of difficulty.
\end{itemize}

\section{Related Work}

\begin{figure}[t]
	\centering
	\includegraphics[width=6cm,trim=0cm 0cm 0cm 0cm, clip]{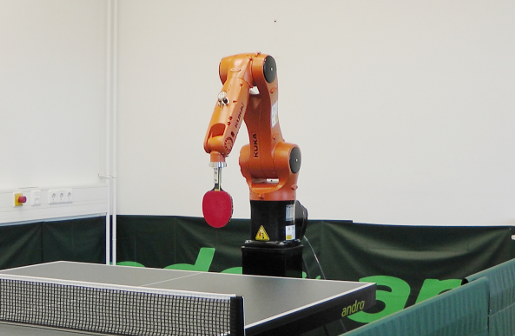}
	\caption{Table tennis robot system with a KUKA Agilus robot. The goal is to learn the orientation and the velocity of the racket at hitting time.}
	\label{fig:robot}
\end{figure}
\subsection{Reinforcement Learning in Robotics}
RL is particularly successful in applications for which information, such as the dynamics, would otherwise be necessary to solve the task \cite{AtariFirst,AlphaGo,RLSurvey}. In those cases thousands of episodes could be generated, which is often not possible in the field of robotics. Different approaches are needed to overcome this drawback. 
Often most of the learning phase is done in simulation and afterwards applied to the real world \cite{peng18_sim2real,chebotar19_sim2real,rusu17_sim2real}. Using multiple robots in parallel can increase efficiency, for example in a door opening task \cite{gu_2017_dooropen} or to collect grasping attempts \cite{levine2018_largescale_robots}. To accelerate learning a difficult task one often includes human knowledge into the RL algorithm. This can mean shaping the reward function \cite{ng99_reward_shaping} directly or including human feedback within the reward signal \cite{Loftin2014_feedback,  saunders18_intervention, christiano2017_preference}. Often expert demonstrations are used for initialization or within training \cite{rajeswaran2018_manip_demonstrations, vecerik2018leveraging, nair2018_exp_demonstrations}. Building upon and improving conventional controllers can make learning in real world scenarios possible \cite{johannink2019_residual_reinforcement, Silver2018ResidualPL}.
The RL-algorithm of this paper is embedded into a robot software environment. This way prior knowledge of the system is utilized to simplify the learning problem. By using data from a prediction algorithm and passing the resulting robot target state to a trajectory planner, we could reduce complexity and learn in very few examples, i.e. playing only $200$ balls with our table tennis robot. 
\subsection{Learning in Robotic Table Tennis}

Robotic table tennis is a challenging field for learning algorithms needing accurate control in a fast-changing noisy environment. Most of the research is done in simulation. \cite{peters2010_policy_search} showed that their Relative Entropy Policy Search method works in a simulated table tennis environment using only a sparse reward. Using a one-step DDPG approach similar to ours \cite{zhu2018_sim_ddpg} could learn very precise policies by simulation up to 200,000 balls. In \cite{gao2020_ttsim} a 8-DOF robot was controlled in joint space with an evolutionary search CNN-based policy training. \cite{Akrour2018_ttsim} developed a trajectory-based monotonic policy optimization and applied it to learning to hit a simulated spinning table tennis ball. \cite{Mahjourian2018} used a virtual reality environment to collect human example strokes and self-train a policy on top of these.

Applying these techniques on a real robot is another challenge and approaches are rarer. The traditional approach to get stroke movements are heuristics \cite{Tebbe2020} or optimization-based trajectory planning \cite{koc20181_traj_gen, Yu2013}. In this context, often a bounce model between ball and racket is utilized \cite{koc20181_traj_gen, omron_paper, nakashima2010, Zhao2016_spinning}. Learning strategies are the alternative. One can train return strokes as a combination of movement primitives \cite{Muelling2013,Peters2013}.  \cite{buechler2020_muscular_arm} even developed a new pneumatic robot arm capable of moving with high accelerations and taught it to smash table tennis balls using PPO. The motion of the opponent's racket is used in \cite{Wang2011} to predict the aim of the opponent and adjust movement timing and generation accordingly.

All these approaches brought promising results, but could only play table tennis in a very limited scenario, such as against a ball throwing machine or using really slow balls. 

\section{The Learning Problem}

Our goal is to teach a KUKA Agilus industrial robot (see. Fig. \ref{fig:robot}) how to play table tennis. Two high-speed cameras are mounted on the ceiling of the robot laboratory to determine the position of the ball. The robot arm is to perform the table tennis stroke in such a way that the ball then hits a target point on the other side of the table with the highest possible precision. An end-to-end learning model using the raw images from the cameras can only be realized with an extremely large number of examples and would need a lot of processing power. We have therefore already developed a tracking system that predicts the trajectory of the ball up to the moment of impact on the bat \cite{Tebbe2018}. As only the point of hitting between ball and racket is essential, we parameterize the stroke movement by position, speed and orientation of the racket at the point of impact with the ball. The position is estimated by our trajectory prediction algorithm. Speed and orientation are outputs of the reinforcement learning problem. Finally, we use them to iteratively plan the arm trajectory using the Reflexxes Library \cite{reflexxes}.

\subsection{Interpretation as a Reinforcement Learning Problem}
Following the usual practice in reinforcement learning, we define our problem as a Markov decision process $(S,A,p,r)$. To reduce complexity, episodes have length $1$, i.e. the transition function $p: S\times A\times S$ maps all states with probability $1$ to the end state $e$. The state space $S = \subset \mathbb{R}^9 \cup \{e\}$ is a $9$-dimensional interval plus end state $e$. Its elements are the vectors concatenating the 3D position, velocity and spin of the table tennis ball just before the stroke. The action space $A = [-20,20]\times[-30,30]\times[0,2]\subset \mathbb{R}^3$ contains elements $(\alpha, \beta, \dot{r}_x)$ consisting of the racket's Euler angles $\alpha, \beta$ in degrees and velocity $\dot{r}_x$ in $\frac{m}{s}$ along the long side of the table at hitting time. Only these values are learned to keep the action space smaller and in practice are supplemented by $\gamma = -0.1 * \dot{b}_y$ and $\dot{r}_y = \dot{r}_z = 0$, where $\dot{b}$ is the velocity of the ball at hitting.

\begin{figure}[t]
	\centering
	\includegraphics[width=0.45\textwidth]{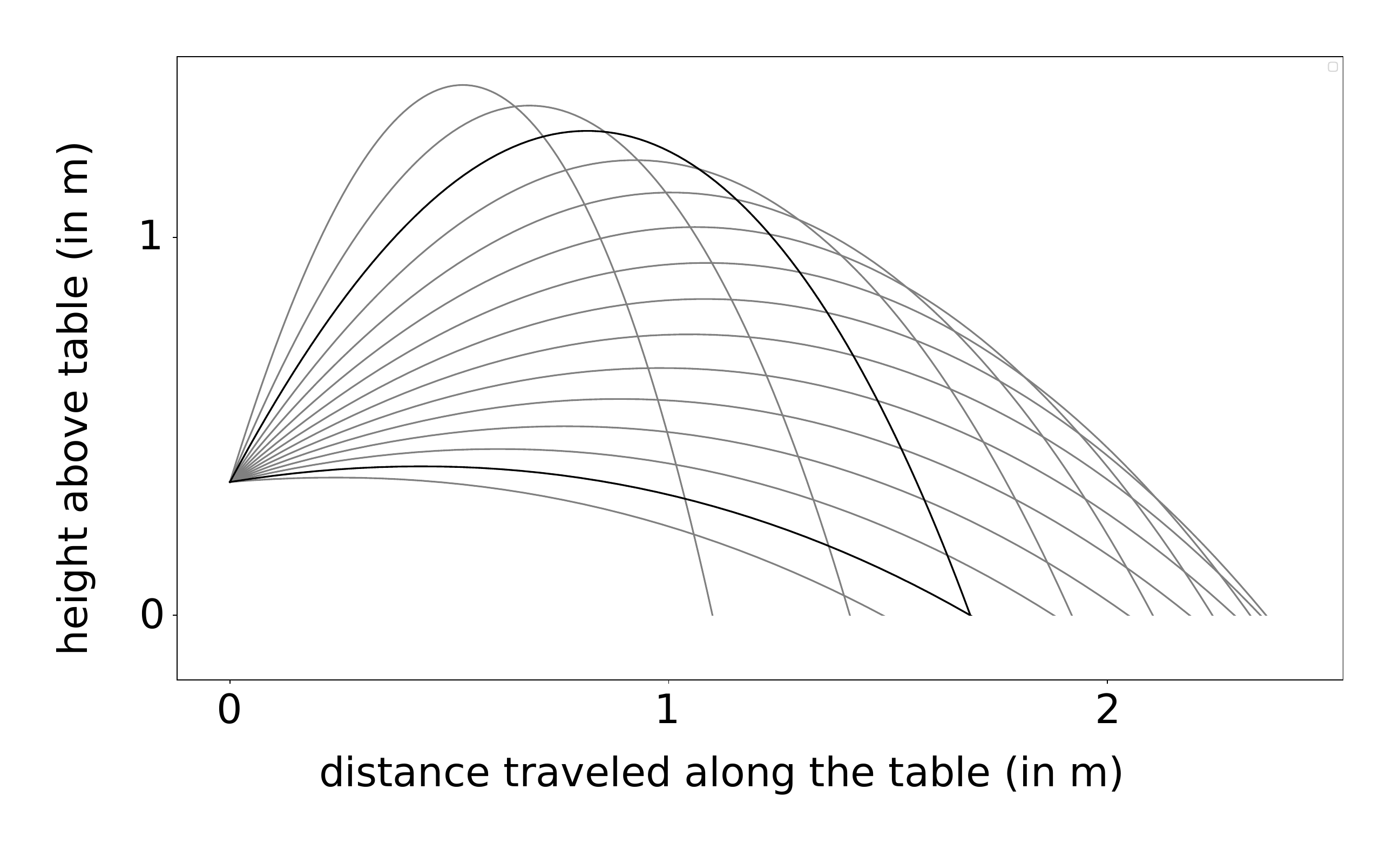}
	\caption{The figure shows several simulated ball tracks with different starting angles, viewed from the side. The ambiguity is evident by the two black trajectories with the same achieved goal position.}
	\label{fig:two_goals}
\end{figure}

\subsection{Reward}
The reward should depend on the distance between achieved goal position and target goal. However, this makes the optimal solution ambiguous. By only changing one angle of the racket orientation we can get two ball trajectories with the same achieved goal reached as illustrated in Figure \ref{fig:two_goals}. One of the trajectories belongs to a very high ball, which is undesirable as it gives the opponent a good opportunity to smash the ball. Thus, we also penalize the height of the ball and define the reward by 
$$R(g_a,h,g_d) = -|g_d - g_a| - \alpha \cdot h$$
where $g_d$ is the desired goal for the landing position of the ball on the table, $g_a$ is the actually achieved goal position, $h$ is the height value of the ball halfway to the goal and $\alpha$ is the coefficient that weights the influence of the height value. We use $\alpha = 0.07$ throughout the experiments, tuned with the simulation from the next section.

\section{The Environments}

\subsection{Simulation}
\begin{figure}[t]
	\centering
	\includegraphics[width=0.48\textwidth]{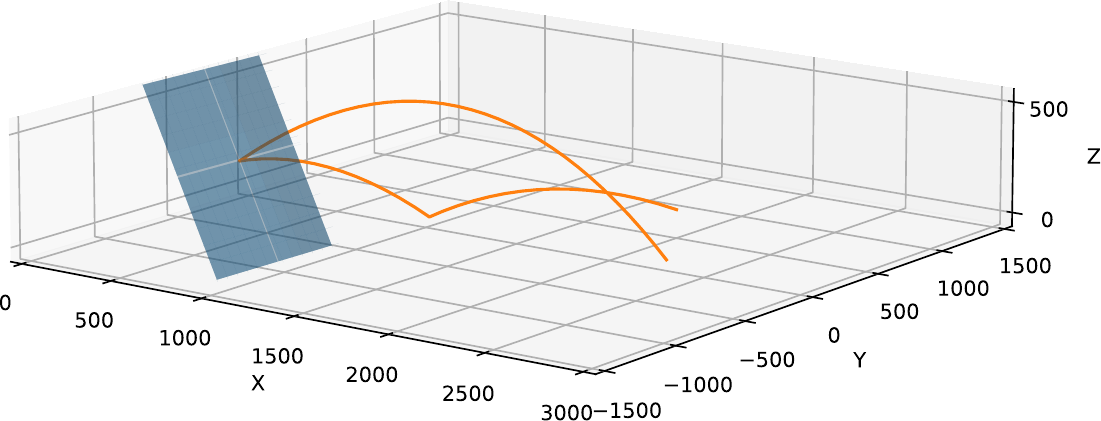}
	\caption{Simulated example trajectory. The racket is represented by the blue plane.}
	\label{fig:sim}
\end{figure}

To verify the functionality of the learning algorithm and for hyperparameter optimization a simulation was designed (see. Fig. \ref{fig:sim}). Note that the simulation is not used to pre-train models for the real robot, but only to design the algorithm and tune hyperparameters. The ball trajectory is calculated by forward solving the following differential equation using a forth order Runge-Kutta method. The underlying equation model \cite{Tebbe2018} is
\begin{align}
\label{eq:accelerations}
\dot{v} = - k_D \left \| v \right \| v + k_M \omega \times v - 
\begin{pmatrix}
0\\ 
0\\ 
g
\end{pmatrix}.
\end{align}
Here $k_D$ is the coefficient for the drag force (air resistance), $k_M$ the coefficient for the Magnus force coming from the angular velocity (spin) of the ball, $g$ is the gravitational constant, and $v (\omega)$ are the translational (angular) velocity of the ball. With this we can estimate the trajectory in midair. 
For the bounce at the table we apply an elastic collision model, where the weight of the ball is negligible in relation to the weight of the table, i.e. $m_b \ll m_t$. In this case we obtain the new $z$ component of the velocity vector by ${v_b}_z' = -{v_b}_z$. Analogously we proceed for the collision between ball and racket. Again, the racket connected to the robot arm is much heavier, $m_b \ll m_r$. First we transform the velocity vectors $v_b$ and $v_r$ so that the $z$ axis is in the direction along the normal of the racket plane. We refer to this transformation as $T$. Then by one-dimensional elastic collision we have $$(Tv_b)'_z = 2*(Tv_r)_z - (Tv_b)_z.$$
While the flight model is rather realistic, the bounce models are now oversimplified. Still, the simulation provides a solid, repeatable test-bed for performance evaluation of the algorithms.

\subsection{Robot}
\begin{figure}[t]
	\centering
	\includegraphics[width=0.49\textwidth,trim=0cm 1cm 0cm 1cm, clip]{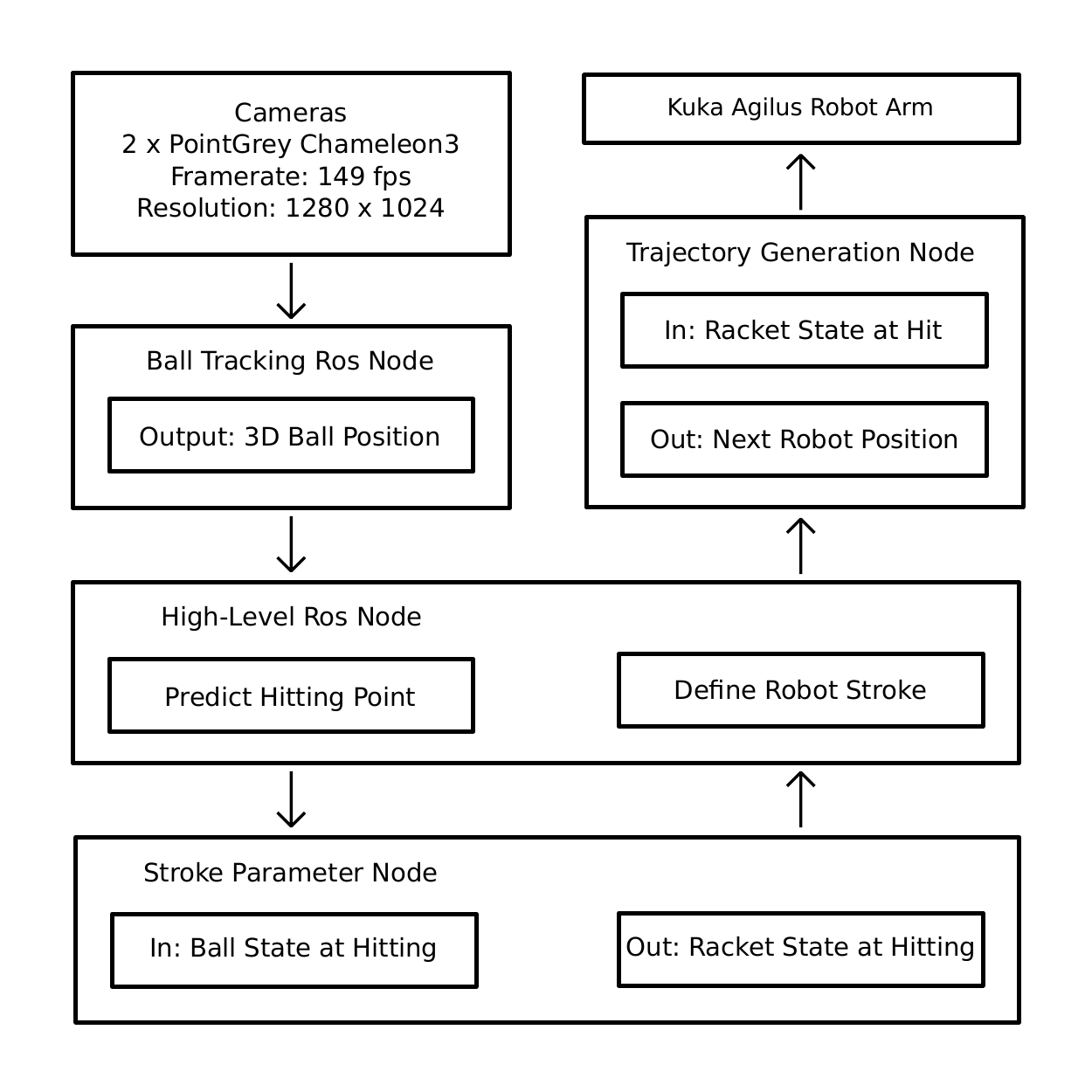}
	\caption{Process on the real robot.}
	\label{fig:ros}
\end{figure}

On the real robot we use the Robot Operating System (ROS). The trained actor network is evaluated in a Python ROS node. The process is illustrated in figure \ref{fig:ros}. We use a stereo system with two PointGrey Chameleon3 cameras to record the table tennis ball. The ball tracking node finds the ball on each camera using traditional image processing and triangulates the pixel positions to output the position of the ball in 3D \cite{Tebbe2018}. In the high-level node the sequence of positions is stored. After an outlier removal the sequence is used to predict the state of the ball at the time it hits the racket. The velocity and the position are estimated using an extended Kalman filter \cite{Tebbe2018}. The spin is derived from the trajectory by using our Magnus force fitting approach \cite{Tebbe2020}. Each new prediction is forwarded to the stroke parameter node where the actor is evaluated. It outputs the desired state of the racket at hitting time which is then sent to the trajectory generation node. Using the Reflexxes library \cite{reflexxes} the trajectory of the robot arm is calculated and finally sent to the KUKA Agilus KR6 R900 robot using the Robot Sensor Interface (RSI).

To give the robot more time for the movement execution we begin with actions computed from early hitting state predictions and gradually refined as more accurate measurements become available. For the purpose of training only the last, most accurate, values are used.

\begin{algorithm}[t]
	\caption{Training algorithm \label{algo:train}}
	\label{dpgalgo}
	\begin{algorithmic}
		\STATE Initialize critic network $Q(s, a | \theta^Q)$ and actor
		$\mu(s,g | \theta^{\mu})$ with weights $\theta^{Q}$ and $\theta^{\mu}$.
		\STATE Initialize replay buffer $B$.
		\STATE Initialize a random process $\mathcal{N}$ for action
		exploration, with large variance in warm-up phase.
		\FOR{episode = 1, E}
		\STATE Receive observation state $s_e$ and desired goal $g_e$.
		\STATE Select action $a_e = \mu(s_e, g_e | \theta^{\mu}) + \mathcal{N}_e$
		according to the current policy and exploration noise.
		\STATE Post-optimize action $a_e$ using the gradient of the reward function $R$:
		\begin{equation*}
		\nabla_{a} R(Q(s, a | \theta^Q),g)|_{s = s_e, a = a_e, g = g_e}
		\end{equation*}
		\STATE Execute action $a_e$ and observe
		reward parameters $r_e$.
		\STATE Store episode $(s_e, a_e,
		r_e)$ in $B$.
		\IF{after warm-up phase}
		\FOR{training\_step = 1, S}
		\STATE Sample a random minibatch of $N$ episodes $(s_i, a_i,
		r_i)$ from the replay buffer $B$.
		\STATE Update critic by minimizing the loss
		\begin{equation*}
		\frac{1}{N} \sum_i \lVert r_i -
		Q(s_i, a_i | \theta^Q)\rVert_2.
		\end{equation*}
		\STATE Update actor policy using the policy gradient
		\begin{equation*}
		\frac{1}{N} \sum_i
		\nabla_{\theta^{\mu}} R(Q(s, \mu(s, g | \theta^\mu)| \theta^Q),g)|_{s = s_i, g = g_i}.
		\end{equation*}
		\ENDFOR
		\ENDIF
		\ENDFOR
	\end{algorithmic}
\end{algorithm}

\section{The Algorithm}

\begin{figure}[t]
	\centering
	\includegraphics[width=0.28\textwidth]{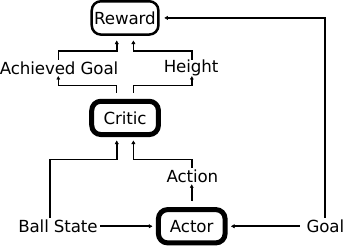}
	\caption{Modified actor-critic model using a parameterized reward.}
	\label{fig:ac}
\end{figure}

Our algorithm uses an actor-critic model similar to DDPG \cite{DDPG} / HER \cite{HER}. The critic is adapted to output a multi-dimensional parameter vector instead of the 1D reward. The vector is used together with the target goal to calculate the corresponding reward. In this way the critic does not need information about the goal, reducing the input dimension of the critic. Using the gradient of the compositon between critic and reward function (see algorithm \ref{algo:train}), the output of the actor is trained to maximize the reward. The approach is depicted in figure \ref{fig:ac} and it will be denoted by APRG (accelerated parametrized-reward gradients).

The deterministic actor-critic model consists of two components. A supervised trained critic network and an actor model outputting the learned policy trained with the help of the critic's gradient (see Fig. \ref{fig:ac}).

\subsection{Critic}
In our scenario the critic receives the ball state (predicted position, velocity and spin at hitting time) and the action (orientation and velocity of the bat) as input and outputs reward parameters (the achieved goal position and average ball height above the table) estimated for the specified state and action. The  $L_2$-loss is used for training. Learning reward parameters and not the reward itself is advantageous. The critic does not need the desired goal as input and the parameters are less complex than the complete reward function, reducing complexity. Also, the outputs are easier to be understood by a human which helps in debugging.

\subsection{Actor}
The actor is fed with the ball state and the target goal position and should return the action. To train the actor we assume the critic weights fixed and use the gradient of the reward with respect to the actor weights. Using this gradient in the optimization step, the actor will use actions which maximize the reward calculated from the critic's output. The training procedure is written down in algorithm \ref{algo:train}.

\subsection{Exploration}
Exploration on the real robot is not suitable for the whole action space. Part of the search space might include robot configurations which are not reachable at all or in the available time. We decided to start recording actions with small Gaussian noise added to a pretested action. With enough samples the gradient of the critic is roughly pointing in the correct direction for improvements, and we can start training. In most cases the actions are then changing in a way using only feasible configurations. 

\subsection{Novelty}
A one-step DDPG approach was already proposed for robotic table tennis \cite{zhu2018_sim_ddpg}. Unfortunately they only showed performance for between $10,000$ and $200,000$ training examples generated in simulation. Our approach is tested also on a real robot and most experiments are conducted on as few as $200$ episodes. It is also differing from classical DDPG by the following modifications:
\begin{itemize}
	\item
	The critic is to output the parameters needed to calculate the reward instead of the reward.
	\item
	We post-optimize the actions via SGD.
	\item We start with a warm-up phase of random actions instead of $\epsilon$-random action in-between learning.
\end{itemize}

\section{Experiments in Simulation}

\begin{figure}[t]
	\centering
	\includegraphics[width=0.45\textwidth]{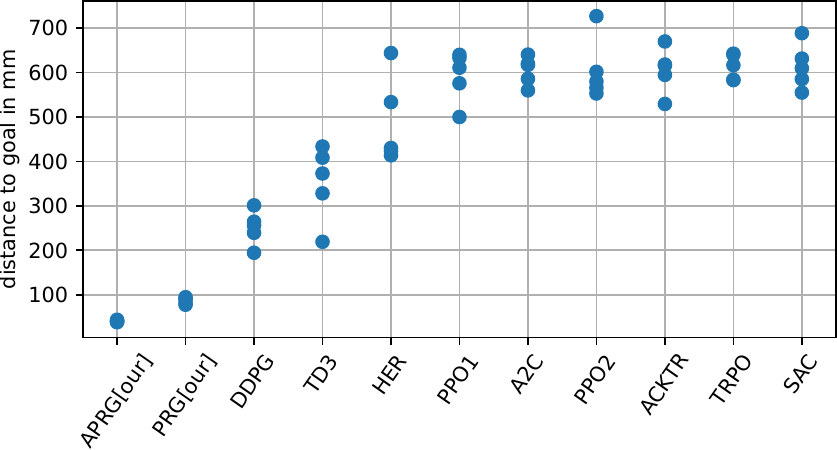}
	\caption{Comparison against the baseline algorithms. Showing the five results for the best parameters from $100$ tested trials using the Optuna hyperparameter optimization framework. }
	\label{fig:sim_result}
\end{figure}

\begin{figure*}[t]
	\centering
	
	\begin{tabular}{cc}
		\includegraphics[width=0.47\textwidth,trim=0cm 0 0cm 0cm, clip]{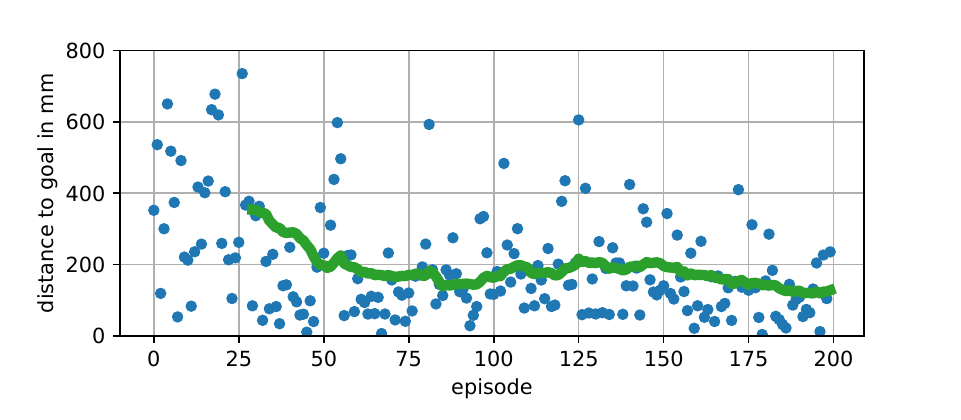}&
		\includegraphics[width=0.47\textwidth,trim=0cm 0 0cm 0cm, clip]{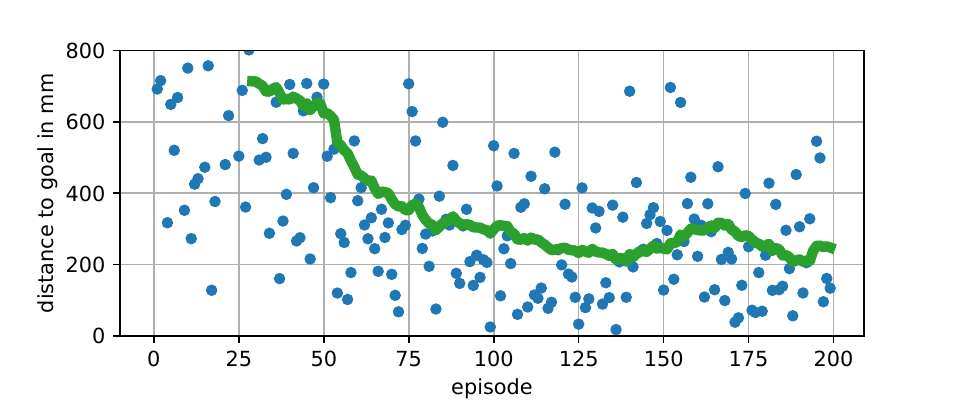}\\
		(a) serve & (b) serve + I-play\\
		\includegraphics[width=0.47\textwidth,trim=0cm 0 0cm 0cm, clip]{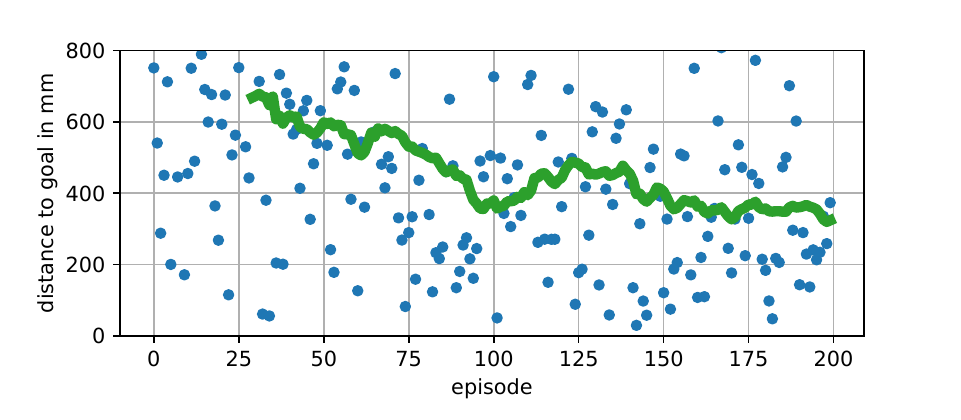}&
		\includegraphics[width=0.47\textwidth,trim=0cm 0 0cm 0cm, clip]{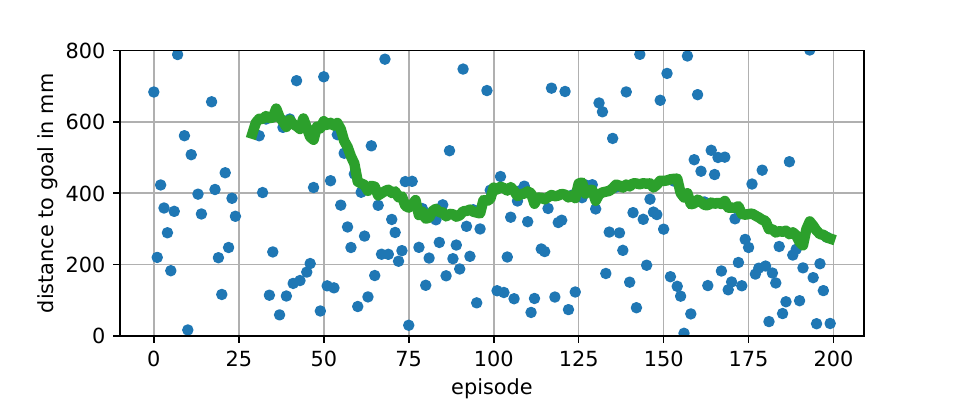}\\
		(c) serve + V-play & (d) serve + X-play\\
	\end{tabular}
	\caption{Result for training on the real robot in four scenarios. The experiments always started with a warm-up phase of $30$ random actions. The green line represents the running average over the $30$ episodes.}
	\label{fig:scenario_result}
\end{figure*}

To get a set of effective hyperparameters we have conducted a parameter search on our simulation with the Optuna framework \cite{Optuna}. For comparison, we also train policies using the state-of-the-art algorithms from the stable baseline repository \cite{stable-baselines} (\cite{A3C,ACKTR,DDPG,HER,PPO,SAC,TD3,TRPO}). 
For a fair evaluation we did hyperparameter optimization for each of the baseline models. As the learning process seems to have a high variance, we decided to average over five tries for each set of parameters. With the Optuna framework we tested $100$ parameter configurations for each method. Figure \ref{fig:sim_result} shows results for the best parameters, respectively. It is evident, that DDPG-based methods (APRG, PRG, DDPG, TD3, and HER) are performing better. 

Our APRG algorithm stays at the top with an average goal error of $40.6mm$. It also shows that post-optimization (APRG) gives better results than unoptimized parametrized reward gradients (PRG), but one can get faster inference time with PRG at the expose of a little accuracy.

For be fair, we must mention that the difference between the algorithms becomes smaller when learning over $2000$ episodes or more. However, performance on a smaller number is more relevant, because in table tennis, rapid adaption plays a major role. In cooperative play with the real robot, human players quickly became impatient when they could not see any improvement in the robot returns.

\section{Experiments on the robot system}
To show that our method also works on the real robot we conducted several experiments of increasing complexity.

\subsection{System noise}
In a first experiment we want to find out how much noise the learning process has to cope with on the real robot. For this purpose we let the ball machine TTMatic 404 serve the ball $200$ times in the same way and let the robot return the ball with unchanging, predefined action parameters. In fact, the balls have the same hitting position with an accuracy of $46.6mm$ and an average speed deviation of $0.92 m/s$. The deviation of the achieved target positions for the resulting robot returns is much larger with an average accuracy of $123.9mm$. We assume that to be the limit achievable in the best case. This shows how challenging the scenario is. 

\subsection{Human play in regular exercises}
In our main experiment the robot is tested against a human player in four increasingly challenging exercise scenarios. The player is playing the ball in a predefined sequence. In this way we can judge the performance for increasing difficulty. These types of exercises are a regular part of table tennis training for amateurs as well as professionals. So a robot capable of learning these could augment human training procedures. The algorithm starts from scratch, i.e. it is not pre-trained by simulation data. The robot begins with a warm-up phase of $30$ random actions and ends with a total of $200$ actions / episodes / ball contacts.

The following scenarios are tested:
\begin{itemize}
	\item Simple backhand serves.
		
	The human is always playing the same serve and the robot has to return to the middle of the table (Goal: $[2000,0]$). 
	
	\item Serve and human I-play.
		
	The human begins with a serve and the rally is continued along the mid-line of the table (Goal: $[2400,0]$).
	
	\item Serve and human V-play.
		
	The human begins with a serve and has to alternate the ball placement between the left and right side of the table, on success forming a V-shape (Goal: $[2400,0]$).
	
	\item Serve and human X-play.
	
	The human begins with a serve and in the following ball exchange the robot and the human place the ball alternately on both sides of the table, forming a X shape if successful (Goals: $[2200,-300]$ and $[2200,300]$).
	
\end{itemize}

The goal coordinates are specified in the coordinate system of the table, where $x=0$ / $1370$ / $2740$ is the robot's table end / net line / human's table end and $y=-762$/$0$/$762$ is the left end / middle line / right end of the table. The odd numbers are coming from the standard table size of $9ft \times 5ft \times 2.5ft$.

The results are presented in figure \ref{fig:scenario_result}. To put the result into perspective, consider the limit of $124mm$ of the fixed action evaluated by the first experiments. So for the serve-only scenario an average of $136mm$ (x: $114mm$, y: $47mm$) to the goal for the last 50 balls is coming very close to that. In the I-play scenario the rally is continued after the serve making it more challenging. A result of $269.2mm$ accuracy in the last 50 episodes is worse, but the x-error is $243.3mm$ and the y-error is $78mm$ showing that the ball is still playing accurately to the middle of the table. The V-play has more deviation from the human player, as each ball is played differently. In this exercise we could achieve a goal error of $329mm$ (x: $177mm$, y: $126mm$). Even more challenging is the X-play achieving a goal error of $393mm$ (x: $282mm$, y: $238mm$). 

\subsection{Human play with different opponents}
The experiments from the last section were all conducted with a player very familiar with the robot and its behavior. To test the robot also against different play styles, we invited three players of the local table tennis club. They were just instructed to play cooperatively with the robot. 

\begin{figure}[h]
	\centering
	\includegraphics[width=0.48\textwidth,trim=0cm 0.5cm 0cm 0.5cm, clip]{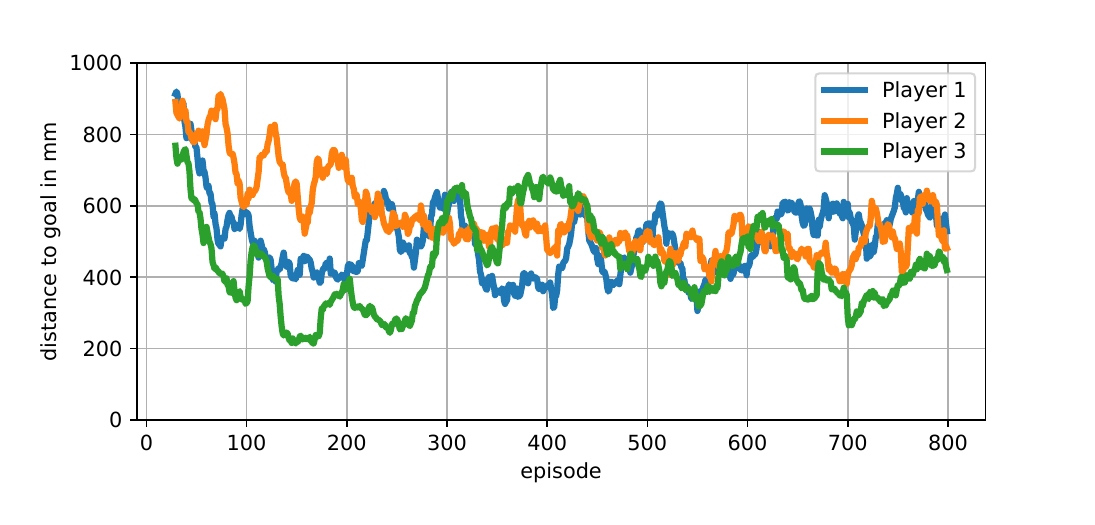}
	\caption{Result for training on the real robot against three human players. The experiments always started with a warm-up phase of $30$ random actions. The lines represent the running average over the last $50$ episodes.}
	\label{fig:players_result}
\end{figure}

Results are presented in figure \ref{fig:players_result}. Performance losses are visible when players have tried new strokes or placements. But the error always converged to an acceptable value.

\subsection{Using a ball throwing machine}
While our main focus is playing against a human opponent, we also did a learning experiment with a ball throwing machine. This scenario is particularly suitable for comparisons, as this is the most common test for table tennis robots in the literature. At first the machine should place the ball only on one spot, analogous to the system noise experiment. The robot should learn the action for returning the ball to the middle of the table (Goal:$[2400,0]$). This results in a very accurate return with a goal error of $118mm$ (x: $85mm$, y: $63mm$) over the last $50$ of $200$ episodes in total. Secondly we change the ball throwing machine to distribute balls evenly on the side of the robot. Here we have an accuracy to the target of $209mm$ (x: $172mm$, y: $88mm$). 

A comparison of our results to other table tennis robots in the literature can be found in table \ref{tab:expRot}. Since most papers only record the return rate of balls successfully played to the opponent half of the table we also included these. It is clearly visible that our approach is achieving state-of-the-art performance. Only \cite{Muelling2013} has a better return rate for an oscillating ball machine.

\begin{table}[h]
	\centering
	\def\arraystretch{1.2}
	\begin{tabular}{>{\centering\arraybackslash}m{2.15cm}|c|c|c|c|c}
		robot model& stroke & balls & hit & return & error  \\
		& type &  & rate & rate & to goal  \\
		\hline
		KUKA Agilus [our] & serve  & 50 & 100\% & 100\% & 135mm\\
		\hline
		[our] & I-play  & 50 & 98\% & 96\% & 269mm\\
		\hline
		[our] & V-play  & 50 & 100\% & 88\% & 329mm\\
		\hline
		[our] & X-play & 50 & 98\% & 92\% & 393mm\\
		\hline
		[our] & occ. BM & 50 & 98\% & 88\% & 209mm\\
		\hline
		[our] & fixed BM& 50 & 98\% & 98\% & 118mm\\
		\hline
		Barrett WAM \cite{koc20181_traj_gen} & occ. BM & 200 & - & 80\% & -\\
		\hline
		Barrett WAM \cite{Muelling2013} & occ. BM & 30 & - & 97\% & 460mm\\
		\hline
		\cite{Muelling2013} & I-play & - & - & 88\% & -\\
		\hline
		Muscular Robot \cite{buechler2020_muscular_arm} & fixed BM & 107 & 96\% & 75\% & 769mm\\
		\hline
		Wu/Kong \cite{Zhao2016_spinning} & occ. BM & 732 & - & 71\% & -\\
		\hline
		lab-made \cite{Zhang2011_tracking} & fixed pos. & - & - & 80\% & -
	\end{tabular}
	\caption{Comparison against other table tennis robots.}
	\label{tab:expRot}
\end{table}

\section{Conclusion and future work}

In this research work a RL algorithm was developed for sample efficient learning in robotics. Extensive experiments were conducted to test it in a real robotic environment. It should determine the parameters for the optimal return of a table tennis ball. The results are measured by the accuracy with respect to a defined target on the table. The learning process is integrated into an existing robot system using a KUKA Agilus KR 6 R900 robot arm. The robot could learn an accurate return in under $200$ balls. This demonstrates robust and effective learning in a very noisy environment. Comparing the success rate of the returns, our algorithm beats the previous research approaches. Beyond the application for robotic table tennis, our method can be used in all cases where the trajectory of a robot can be represented by a lower-dimensional parameter vector, as in our case orientation and speed at the hitting point. 

On the way to competitive play against top human players there is still a lot to do. In the future we plan to let our robot learn in even more challenging match realistic scenarios. This requires generalization for many more domains like serve/no serve, topspin/backspin/sidespin, short/long balls etc. The goal parameters should also include speed and spin, which will be needed for a successful strategy capable of beating advanced human players.

\addtolength{\textheight}{-5.2cm}   




\bibliographystyle{IEEEtran}
\bibliography{root}

\end{document}